\documentclass[twocolumn]{autart}    % Enable this line and disable the 
\usepackage{graphicx}          % Include this line if your 
\usepackage{hyperref}
\usepackage{algorithmic}
\usepackage{algorithm}
\usepackage{array}
\usepackage{bm}
\usepackage{amsmath}
\usepackage{amssymb}
\usepackage[utf8]{inputenc}
\usepackage{cite}
\usepackage{color}

\newtheorem{assumption}{Assumption}
\newtheorem{definition}{Definition}
\newtheorem{lemma}{Lemma}
\newtheorem{problem}{Problem}
\newtheorem{remark}{Remark}
\newtheorem{theorem}{Theorem}

\def\b{\boldsymbol}
\def\be{\boldsymbol e}

\def\bu{\boldsymbol u}
\def\bv{\boldsymbol v}

\def\bg{\boldsymbol g}

\def\bp{\boldsymbol p}
\def\bfx{\mathbf x}
\def\R{\mathbb R}
\def\Z{\mathbb Z}
\def\h{\mathrm h}
\def\bg{\mathbf g}
\def\P{\mathbf P}
\newcommand{\V}{\mathbf V}
\newcommand{\E}{\mathbf E}
\newcommand{\Var}{\mathbf{Var}}
\newcommand{\trace}{\mathbf{trace}}

\newcommand{\fr}[1]{\texttt{#1}}
\def\fW{\fr{W}}

\def\a{\alpha}
\def\B{\beta}
\def\D{\Delta}
\def\bX{\bm{X}}
\def\bbX{\bar{\bm{X}}}
\def\bbp{\bar{\bm{p}}}
\def\bbv{\bar{\bm{v}}}
\def\bbu{\bar{\bm{u}}}
\def\bq{{\bm{q}}}
\def\bzeta{\bm{\zeta}}
\newcommand{\Expect}[1]{\mathbf{E}\left(#1\right)}

\begin{document}

\begin{frontmatter}
                                              % is not shown in the output.

\title{Aerial Target Encirclement and Interception with Noisy Range Observations\thanksref{footnoteinfo}} % Title, preferably not more 

\thanks[footnoteinfo]{This research was supported by National Research Foundation of Singapore under its Medium-Sized Center for Advanced Robotics Technology Innovation, National Natural Science Foundation of China under Grants U21A20476. Corresponding author:
Shanghai Yuan.}

\author[Paestum]{Fen Liu}\ead{fen.liu@ntu.edu.sg}, 
\author[Paestum]{Shenghai Yuan}\ead{shyuan@ntu.edu.sg},
\author[Paestum]{Thien-Minh Nguyen}\ead{thienminh.nguyen@ntu.edu.sg},
\author[Rome]{Wei Meng}\ead{meng0025@ntu.edu.sg},   
\author[Paestum]{Lihua Xie}\ead{elhxie@ntu.edu.sg},% Add the 

\address[Paestum]{School of Electrical and Electronic Engineering, Nanyang Technological University, Singapore 639798, Singapore}  % Please supply                                              
\address[Rome]{School of Automation, Guangdong University of Technology, Guangzhou 510006, China}             % full addresses
%\address[Baiae]{The White House, Baiae}        % here.

                             % keyword list or with the 
                                          % help of the Automatica 
                                          % keyword wizard

\begin{abstract}                          % Abstract of not more than 200 words.

This paper proposes a strategy to encircle and intercept a non-cooperative aerial point-mass moving target by leveraging noisy range measurements for state estimation. In this approach, the guardians actively ensure the observability of the target by using an anti-synchronization (AS), 3D ``vibrating string" trajectory, which enables rapid position and velocity estimation based on the Kalman filter. Additionally, a novel anti-target controller is designed for the guardians to enable adaptive transitions from encircling a protected target to encircling, intercepting, and neutralizing a hostile target, taking into consideration the input constraints of the guardians. Based on the guaranteed uniform observability, the exponentially bounded stability of the state estimation error and the convergence of the encirclement error are rigorously analyzed. Simulation results and real-world UAV experiments are presented to further validate the effectiveness of the system design.

\end{abstract}

\end{frontmatter}

\section{Introduction}
The growing accessibility of micro aerial vehicles (MAVs) has introduced critical security risks. MAVs can breach restricted airspace, including airports, confinement centers, and government facilities, causing significant damage due to the limited effectiveness of existing countermeasures \cite{zheng2025optimal,li2022three}.
%Traditional aerial target countermeasure systems, such as infrared and radar-based defenses, rely heavily on a direct line of sight (LOS) \cite{yuan2024MMAUD}. 
Traditional aerial defense systems, like bulky LiDAR and ground radar, face deployment challenges due to their size, weight, and operational constraints \cite{yuan2024MMAUD}.
Alternative methods like radio frequency jamming, GPS spoofing, high-power lasers, and net-based interception \cite{souli2023multi} are constrained by energy demands, limited flexibility, and operational complexity, reducing their effectiveness in dynamic scenarios.
To address these challenges, modern solutions use autonomous MAV swarms for scalable and adaptive countermeasures, offering coordinated detection and interception. However, practical implementation must carefully consider critical factors such as target localization, payload constraints, and computational limitations.
\vspace{-3mm}

Existing target localization algorithms using single sensors mainly rely on cameras for bearing or angle estimation \cite{ chen2021observer, chen2022triangular}.
Distance measurement is often more straightforward than angle or bearing measurement, as angle estimation requires a complex visual processing pipeline \cite{xu2025airslam}. However, some range-based perception systems \cite{fang2023distributed}, such as UWB or WiFi, need cooperative targets, making them less versatile.
For hostile non-cooperative targets, newer methods provide effective range solutions using Received Signal Strength Indicators (RSSI) \cite{jia2018received}. However, range-based localization algorithms face inherent limitations. These include the requirement for the ranging network to satisfy the analogous range rigidity theory \cite{fang2023distributed,tang2022relaxed} or reliance on multi-step ego motion to observe the target \cite{nguyen2019persistently,jiang20193}. This requires a sufficient number of followers simultaneously tracking the target, or specific conditions such as the target being stationary or moving at low or predictable speeds \cite{shames2011circumnavigation,dong2020target}. When range measurements are insufficient or noisy, general theoretical results addressing targets moving at unknown or variable velocities remain largely unexplored.

Although advances in range-based localization techniques have improved target tracking, integrating them into effective airborne interception systems remains a challenge. 
Approaches using deep reinforcement learning (DRL) \cite{li2023predator} require extensive training data and computational power, making real-time deployment impractical. Furthermore, most control techniques focus on detection and tracking \cite{nguyen2019persistently} without enabling direct engagement, limiting their effectiveness against agile and evasive targets.
Innovative, resource-efficient frameworks are needed to bridge these gaps, ensuring precise target estimation and real-time interception in dynamic environments.

\vspace{-7pt}
To address existing gaps, this paper explores the idea of utilizing guardians to provide overwatch, intercept, and incapacitate unauthorized or malicious targets without continuous LOS ground guidance. This work, to the best of our knowledge, is the first to showcase autonomous 3D moving non-cooperative target interception using minimal onboard sensors with noisy range measurements. The key contributions of this paper are summarized as follows. 

\vspace{-7pt}
Firstly, the position and velocity of a \textit{non-cooperative 3D hostile target} are estimated by using noisy distance measurements from the onboard sensors of two guardians. By transforming the distances into linear observations related to the target state, the Kalman filter can be applied. Compared to existing state estimation works \cite{tang2022relaxed,haring2020stability}, the uniform observability of the hostile target under limited noisy measurements is rigorously proved.

\vspace{-7pt}
Secondly, we propose a novel anti-target controller that integrates the anti-synchronization (AS) encirclement strategy \cite{liu2023multiple, liu2023non} with a 3D trajectory of a vibrating string generated by combining vertical and 2D horizontal encirclement motions. Unlike existing convoy protection \cite{dou2020target,hu2023cooperative}, surveillance \cite{chen2021observer,jiang20193}, and
interception \cite{souli2023multi,shah2014guidance,zheng2021time} works, our proposed control strategy enables adaptive encirclement transitions between protecting a target and confronting a hostile one. It accounts for the physical flight speed limits of the guardians, establishes persistent excitation (PE) of the relative position between the two guardians, and ensures the uniform observability of the hostile target.
\vspace{-2mm}

Finally, we demonstrate the effectiveness of the proposed framework in both simulation and real-world experiments. \vspace{-2mm}

The article is organized as follows: Section 2 introduces the system model and formulates the problem. Section 3 presents the design of the estimator and controller, along with their convergence analysis. Section 4 validates the theoretical findings through numerical simulations and physical experiments. Finally, Section 5 summarizes the key conclusions.
\vspace{-2mm}

Notations: We denote $\R^{+}$ and $\Z^{+}$ as sets of positive real numbers and positive integers. The eigenvalue of matrix $A$ is denoted as $\lambda \{A\}$, where the largest and smallest eigenvalues are $\lambda_{\max}\{A\}$ and $\lambda_{\min}\{A\}$, respectively. The transpose of a matrix $A$ is represented by $A^\top $, and $A^{-1}$ denotes its inverse. We denote the $n\times n$ identity and zero matrices as $I_n$ and $\bm{0}$, and unless specified, $I$ and $\bm{0}$ denote $I_3$ and $\bm{0}_3$. The Euclidean norm is denoted as $||\cdot||$.
For a time-varying quantity $\bm{X}(k)$, we may omit the time step $(k)$ to simplify the notation.
The mean and variance of a random variable $\bX$ are denoted by $\Expect{\bX}$ and $\Var\{\bX\} \triangleq \Expect{\bX\bX^\top}$, respectively. \vspace{-2mm}

\vspace{-2mm}\section{System Model and Problem Formulation}\vspace{-2mm}
\subsection{System Model}\vspace{-2mm}
This study focuses on protecting a target (person, drone, or vehicle) from the threats posed by a 3D non-cooperative hostile target (capable of stable and controlled flight). To accomplish this task, we deploy two MAV guardians, designated guardian 1 and guardian 2. We assume that the attitude changes of guardians are small so that the model can be approximated as a point mass.\vspace{-2mm}

For guardian $i$, $i \in \{1, 2\}$, we define $\bm{X}_i \triangleq (\bp_i,\bv_i)\in \R^{6}$ as its state vector, where $\bp_i \in \R^3$ is the position and $\bv_i \in \R^3 $ is the velocity w.r.t. a common inertial frame $\fW$. We define $\bu_i(k) \in \R^3$ as the acceleration control input, and $t$ is the sampling period. The motion model of guardian 1 and guardian 2 can therefore be stated as follows \cite{li2022three}:\vspace{-2mm}
\begin{equation}\label{eq: dynamic}
\begin{split}
\bm{X}_i(k+1)=A\bm{X}_i(k)+B\bu_i(k), \ i \in \{1,2\},
\end{split}
\end{equation}
where
{
$
A=
\begin{bmatrix}
    I      &tI \\
    \bm{0} & I \\
\end{bmatrix}
,
B=
\begin{bmatrix}
\frac{1}{2}t^2I\\
t I \\
\end{bmatrix}.
$}

\vspace{-1mm}
To better differentiate the targets, the protected target is designated as Target 1, and the hostile target as Target 2.
Let the state of the target $j$, $j\in\{1, 2\}$ be $\bbX_j =(\bbp_j, \bbv_j)\in \R^{6}$ with the position $\bbp_j\in \R^3$ and the velocity $\bbv_j\in \R^3$, both in the frame $\fW$. Similarly, we define the motion model of the targets as \vspace{-2mm}
\begin{equation}\label{eq2}
\begin{split}
\bbX_j(k+1)=&A\bbX_j(k)+B\bbu_j(k), \ j \in \{1, 2\},
\end{split}
\end{equation}
where $\bbu_j(k) \in \R^3$ is the stochastic acceleration, which includes unknown disturbances such as maneuvering. We assume that the target's acceleration follows a heavy-tailed distribution \cite{chen2017maximum} given by,\vspace{-2mm}
\begin{equation*}
\begin{split}
&\bbu_j
\sim
\pi_j\mathcal{N}(0,\bm{W}_{1,j})
+
(1-\pi_j)\mathcal{N}(0,\bm{W}_{2,j}),\\
&\P(\pi_j)=\gamma_j^{\pi_j}(1-\gamma_j)^{(1-\pi_j)},\ \pi_j \in \{0, 1\},\ \gamma_j \in \R^{+},
\end{split}
\end{equation*}
%\vspace{-2pt}
where $\pi_j$ is the mixing weight and satisfies Bernoulli distribution, $\gamma_j$ is the occurrence probability and $\bm{W}_{1,j}$ and $\bm{W}_{2, j}$ are the variances. This distribution satisfies $\gamma_j\gg1-\gamma_j$ and $\bm{W}_{1,j}\ll \bm{W}_{2,j}$.
The acceleration profile models a target that usually moves by moderate exertion but can occasionally accelerate aggressively to escape encirclement or other threats. \vspace{-2mm}

Hence, we define the relative position between the guardians or targets as follows:\vspace{-2mm}
\begin{align}\label{eq3}
    \bq_{ij} \triangleq \bp_i - \bp_j,\
    \bq_{i}^{j} \triangleq \bp_i - \bbp_j,\
    \bq^{ij} \triangleq \bbp_i - \bbp_j,
\end{align}
where the subscript is reserved for the index of a guardian and the superscript for a target. Similarly, the relative distances between the guardians or targets are denoted as follows
\begin{align}
    d_{ij} \triangleq \|\bq_{ij}\|,\ d_{i}^j \triangleq \|\bq_{i}^j\|,\ d^{ij} \triangleq \|\bq^{ij}\|.
\end{align}
\vspace{-0.75cm}

In this work, we assume that the protected target (Target 1) will share its state $\bbX_1$ with both the guardians.
The hostile target (Target 2) is non-cooperative, i.e. the state $\bbX_2$ cannot be directly obtained by the guardians. Therefore, we denote $\hat{\bbX}_2=(\hat{\bbp}_2, \hat{\bbv}_2)$ and $\hat{d}^{12}$ respectively as the estimate of $\bbX_2$ and ${d}^{12}$, which will be furnished via some estimation laws. Furthermore, we define three decision zones $\hat{d}^{12}$, namely $\Omega_1=\{\hat{d}^{12}|\hat{d}^{12}\geq l_3\}$, $\Omega_2=\{\hat{d}^{12}|l_3> \hat{d}^{12}\geq l_2\}$ and $\Omega_3=\{\hat{d}^{12}|l_2>\hat{d}^{12}\geq l_1\}$, where $l_3$, $l_2$ and $l_1$ are the minimal radii of the \textit{target protection zone},  \textit{warning zone} and \textit{take down (or capture) zone}, respectively, and $l_3 > l_2 > l_1$.
In addition, we define the \textit{encirclement shape} vector as \vspace{-2mm}
\begin{equation}
\begin{split}
\bzeta(k)
=
r(k)
\begin{bmatrix}
    \sin(\rho k\pi), &\cos(\rho k\pi), &\h(k)
\end{bmatrix}^\top,
\end{split}
\end{equation}
where $r(k)$ is the encirclement radius, and $\h(k) \in \R$ is the vertical motion function, which can be user-defined. The term \textit{encirclement shape} is chosen by the fact that $\bzeta(k)$ is a time-varying vector that will chart out a certain shape to achieve the desired encirclement. In addition, we assume that $r(k)$ and $h(k)$ are chosen so that $\exists \bar{r} \in \R^{+}$ so that $\|\bzeta(k)\|\leq \bar{r},\ \forall k$, and the sequence $\{\bzeta(k)\}$ satisfies the following PE assumption. \vspace{-2mm}

\begin{assumption}\label{Persistently_exciting_shape}
The sequence $\{\bzeta(k)\}$ is PE, i.e. there exist $\hat{a}_\zeta, \check{a}_\zeta \in \R^{+} $ and $N \in \Z^+$ such that: \vspace{-2mm}
\begin{equation*}
\hat{a}_\zeta I
\leq
S \triangleq\sum_{m=k}^{k+N-1}\bzeta(m) \bzeta(m)^\top\\
\leq \check{a}_\zeta I , \forall k \geq 0.
\end{equation*}
% where $I$ is the $3$-dimensional identity matrix.
\end{assumption}
\vspace{-0.5cm}

The assumption above can be easily satisfied by designing a cyclic sequence of $\bzeta(k)$ vectors that span $\R^3$ with period $N$, and the constants $\hat{a}_\zeta$, $\check{a}_\zeta$ can be specified by directly computing the singular value decomposition of $S$ \cite{klema1980singular}. With the PE property guaranteed, the following definitions are provided.\vspace{-2mm}

\begin{definition} \label{AS_manner}
\textit{The guardian are said to encircle Target $j$ in an AS manner if the preset encirclement shape $\bzeta$ and the relative positions $\bq_{i}^{j}$, $i, j \in \{1, 2\}$ satisfy $\bzeta^\top\bq_{1}^{j}=-\|\bzeta\|\|\bq^j_1\|$ and $\bzeta^\top\bq_{2}^{j}=\|\bzeta\|\|\bq^j_2\|$.}
\end{definition}\vspace{-2mm}

\begin{definition} \label{taken_down_radius}
\textit{The hostile target is captured (or taken down) when the radius of the encirclement is sufficiently small, i.e. $r(k)\leq r_c$ with given constant $r_c$.}
\end{definition}\vspace{-2mm}

\begin{figure}
\centering
  \includegraphics[width=0.8\linewidth]{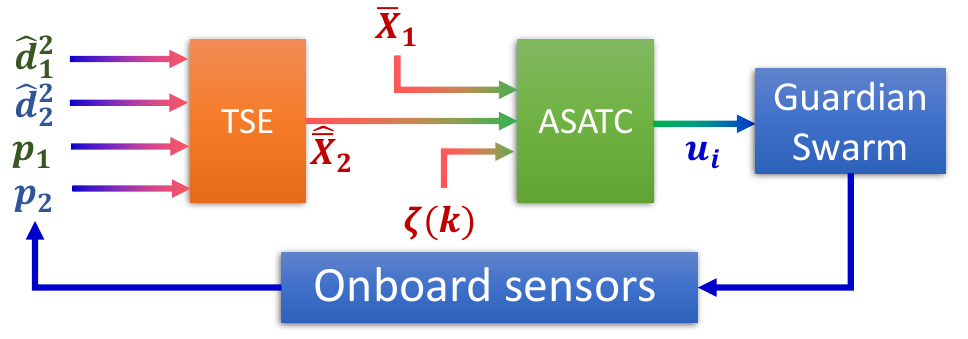}
 \caption{System overview. The Target State Estimator (TSE) is described in detail in Sec. \ref{sec: tpe}. The Anti-synchronization based Anti-target Controller (ASATC) is presented in Sec. \ref{sec: asads}.
 }
  \label{sysflow}
\end{figure}

\subsection{Problem formulation}\vspace{-2mm} \label{sec: problem fomulation}
The overview of the system can be seen in Fig. \ref{sysflow}. In this work, the distance $d_{i}^2$ can be measured by the guardian onboard sensors $i$, e.g., by the equirectangular perception module, RSSI \cite{jia2018received}, or the sound noise strength of the target captured by a microphone \cite{yang2023av}. According to equations (1) and (2) in \cite{jia2018received}, the squared distance $(d_{i}^2)^2$ can be measured from the received power $a^2$ for some known types of targets. 
In theory, $(d_{i}^2)^2$ can be estimated from multiple samples $a^2_s,\ s\in\{1,2\ldots\}$ from multiple wavelengths in the least-squares manner.

Hence, we assume that the actual measured squared distance $(\hat{d}_{i}^2)^2$ differs from the true distance $d_{i}^2$ by an additive error, i.e. $(\hat{d}_{i}^2)^2=(d_{i}^2)^2 + \eta_{i}$, where $\eta_{i}$ is the measurement noise of the guardian $i$ and satisfies
$\eta_{i} \sim \mathcal{N}(0,\sigma_i),\ \sigma_i \in \R^{+}$. Here, $\eta_{1}$ and $\eta_{2}$ are mutually independent.\vspace{-2mm}

Moreover, there is communication between the two guardians. 
We also assume that guardians can align their local frame of reference to $\fW$ and can update their positions $\bp_i$ in real-time by some onboard self-localization system, such as visual-inertial odometry (VIO) or optical flow. In this work, we do not consider the accumulated errors from the onboard self-localization system, which can be mitigated by loop closures or some other techniques \cite{esfahani2019orinet}.\vspace{-2mm}

\begin{assumption}\label{variance bound}
The variances of target acceleration $\bbu_2$ and measurement noise $\eta_i$ are assumed to be positive definite with positive upper and lower bounds, i.e. 
\begin{align*} 
\hat{w}_iI
&\leq \bm{W}_{i,2}\leq\check{w}_iI,
\\
\hat{\sigma}\leq&\sigma_i\leq\check{\sigma}, \ i \in \{1, 2\},
\end{align*}
where $\hat{w}_i$, $\check{w}_i$, $\hat{\sigma}$ and $\check{\sigma}$ all are positive numbers.
\end{assumption}\vspace{-2mm}

Now, we can state the problem of our interest as follows:\vspace{-3mm}
\begin{problem} \label{problem}
Design a range-based estimation law $\hat{\bbX}_2$ for the position of the hostile target. Simultaneously, design a control law $\b{u}_i$ for the guardian to encircle the protected target when $\hat{d}^{12} \in \Omega_1$, and \textit{encircle} or \textit{take down} the hostile target when $\hat{d}^{12}$ enters $\Omega_2$ and $\Omega_3$, respectively. Our objectives are to design estimation and control laws under which there exist finite constants $\hat{\varepsilon}, \varepsilon_1, \varepsilon_2 \in \R^{+}$ and $K \in \Z^{+}$ such that the following can be achieved: $\forall i \in \{1, 2\}$,\vspace{-2mm}
\begin{align}
&\Expect{\|\bbX_2-\hat{\bbX}_2\|^2}
    \leq
\hat{\varepsilon},
\forall k > K,
\label{eq5-1}\\
&\Expect{\|\b{q}_{1}^{1}+\b{q}_{2}^{1}\|^2}
    \leq
\varepsilon_{1}, \ k\in\{k|\hat{d}^{12} \in \Omega_1\},\
\label{eq5-2}\\
&\Expect{\|\b{q}_{1}^{2}+\b{q}_{2}^{2}\|^2}
    \leq
\varepsilon_{2},\ k
\in
\{k|\hat{d}^{12} \in (\Omega_2 \cup \Omega_3)\},\label{eq5-3} 
\end{align}
where $\hat{\varepsilon}$ is an estimation error bound for the hostile target, and $\varepsilon_{1}, \varepsilon_{2}$ are the AS-based encirclement error bounds for the protected target or the hostile target, which can be proven by the theorems presented later.\vspace{-2mm}

In Problem \ref{problem}, the formulas \eqref{eq5-2} and \eqref{eq5-3} imply that the guardians can achieve the encirclement of the protected target or the hostile target in an AS manner stated in Definition \ref{AS_manner}. In other words, the positions of the guardian from the target are opposite to each other. According to Definition \ref{AS_manner} and Assumption \ref{Persistently_exciting_shape}, when $\bzeta(k)$ is PE, we can ensure that the relative position $\bq_{12}$ is also PE, which ensures the observability of the target, as detailed in Lemmas \ref{lem: position_Persistently_exciting} and \ref{Uniform Observability_1} later.
Moreover, the switching between encirclement and capture is based on the radius $r(k)$. Based on Definition \ref{taken_down_radius}, when $r(k) \leq r_c$, it means that the target has been captured or taken down. Note that for ground targets, we only ensure that the two guardians achieve encirclement in the XY plane. 
\end{problem}\vspace{-2mm}

\section{Estimator and Controller Design}\vspace{-2mm}

\subsection{Range-based State Estimation} \label{sec: tpe}\vspace{-3mm}
% Based on the distances $d_{i}^2$ and the positions $\bp_i$, the target $j$ may be located at any point on the sphere with center $\bp_i$ and radius $d_{i}^2$. Considering the presence of two guardians, the target's position can be further constrained to the circle where the two spheres intersect, as illustrated in Fig. \ref{target_motion}. When the actual measurements $(\hat{d}_{i}^2)^2$ include noise, the target's actual motion will deviate from the one shown in Fig. \ref{target_motion}.\vspace{-2mm}

\begin{figure}
\centering
\includegraphics[width=0.8\linewidth]{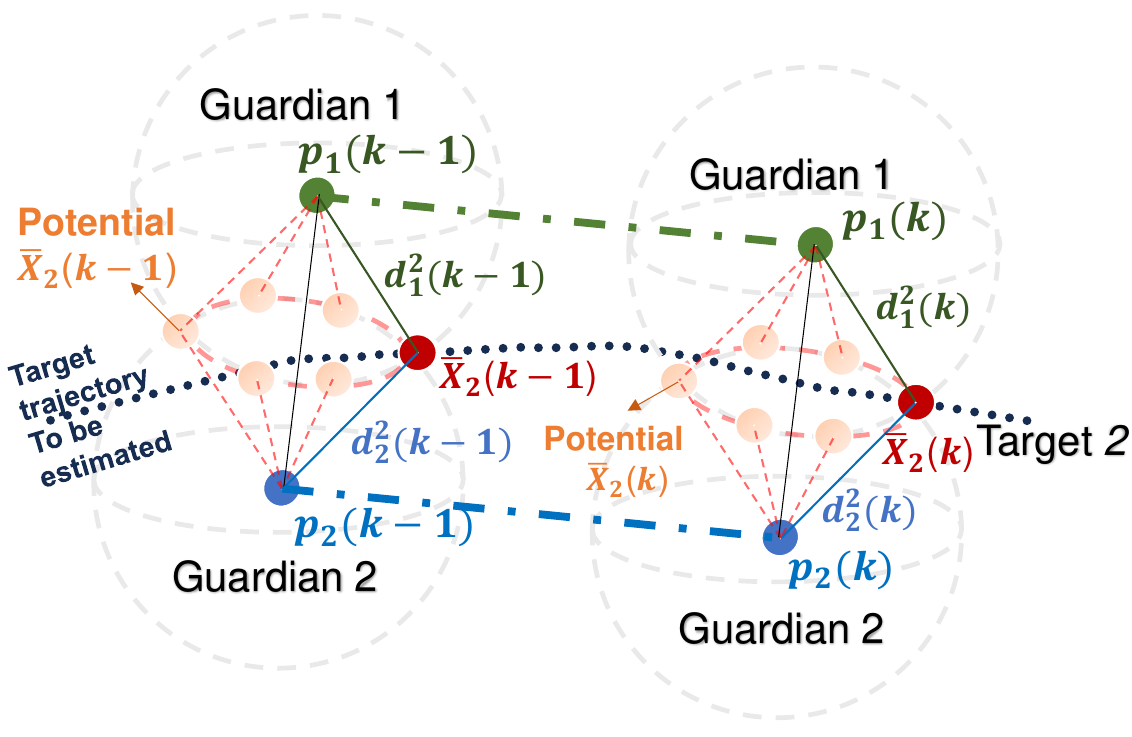}
\vspace{-0.75cm}
\caption{Range-based target motion analysis without measurement noise.}\label{target_motion}
\end{figure}

As shown in Fig. \ref{target_motion}, target 2 is located at any point on the sphere with center $\bp_i$ and radius $d_{i}^2$, assuming that there is no distance noise. Based on noisy measurements $(\hat{d}_{i}^2)^2$ and the fact that $(d_{i}^2)^2=\bp_i^\top\bp_i-2\bp_i^\top\bbp_2+\bbp_2^\top\bbp_2$, we can obtain a stochastic observation $Y \in \R$ that is linearly related to $\bbX_2$ as follows:\vspace{-2mm}
\begin{equation}\label{eq8}
\begin{split}
Y
&=
-\frac{1}{2}\left((\hat{d}_{1}^2)^2-(\hat{d}_{2}^2)^2-\bp_1^\top\bp_1+\bp_2^\top\bp_2\right)
\\
&\triangleq C(k)\bbX_2+\bar{\eta},
\end{split}
\end{equation}
where $C(k)\triangleq\bq_{12}^\top [I\ \bm{0}] \in \R^6$, and $\bar{\eta}=-\frac{1}{2}(\eta_1-\eta_2)$. From $\eta_i \sim \mathcal{N}(0, \sigma_i)$, we have $\Expect{\bar{\eta}}=0$ and the variance $\Var\{\bar{\eta}\}=1/4(\sigma_{1}+\sigma_{2})\triangleq\Gamma_2$.\vspace{-2mm}

Recalling the dynamic model $\eqref{eq2}$ and based on the output $\eqref{eq8}$, the target state estimator (TSE) is designed based on the Kalman Filter in \cite{reif1999stochastic} for a linear time varying (LTV) system as follows:\vspace{-2mm}
\begin{align}
&\hat{\bbX}_2(k)
=A\hat{\bbX}_2(k-1)+\bm{K}(k)\left[Y(k) -\hat{Y}(k)\right],\label{eq9-1}
\\
&\hat{Y}(k)
= C(k)A\hat{\bbX}_2(k-1),\label{eq9-2}
\\
&\bm{K}(k)
=
\bm{G}(k|k-1)C(k)^\top \big[C(k)\bm{G}(k|k-1)C(k)^\top \nonumber\\
&~~~~~~~~~~~+\Gamma_2\big]^{-1},\label{eq9-3}
\\
&\bm{G}(k|k-1)
=A\bm{G}(k-1)A^\top +BW_2 B^\top,\label{eq9-4}
\end{align}
where $\bm{K}$ is the estimator gain, $\bm{G}(k) \triangleq \Var\{\bbX_2-\hat{\bbX}_2\}$ is the estimation error variance, and $\bm{G}(k|k-1)\triangleq\Var\{\bbX_2-A\hat{\bbX}_2(k-1)\}$ is the predicted variance.
Furthermore, denote the variance of the target acceleration $\bbu_2$ as $W_2\triangleq\Var\{\bbu_2\}$. Recalling the distribution of $\bbu_2$ in Sec. \ref{sec: problem fomulation}, we have $W_2=\gamma_i \bm{W}_{1,2}+(1-\gamma_i)\bm{W}_{2, 2}$.\vspace{-2mm} 

\begin{remark}
    In the case when the global positions $\bp_1$ and $\bp_2$ are not available but the local relative position $\bq_{12}$ can be measured, we can define $Y_i, i\in\{1,2\}$ that is related to the relative state $\bbX_{2,i}:=[\bq_i^2,\bv_i-\bbv_2]^\top$ as \vspace{-2mm}
\begin{equation*}
\begin{split}
    Y_i =& (-1)^i \frac{1}{2}\left((\hat{d}_{\nu}^2)^2-(\hat{d}_{i}^2)^2-\bq_{12}^T\bq_{12}\right)\\
    \triangleq& C(k)\bbX_{2,i}+\bar{\eta}_i,
\end{split}
\end{equation*}
where $\bar{\eta}_i=(-1)^i\frac{1}{2}(\eta_\nu-\eta_i)$ with $\nu \neq i \in \{1,2\}$. In this case, we can design the update relative state estimator law as \vspace{-2mm}
\begin{equation*}
\hat{\bbX}_{2,i}(k)
=A\hat{\bbX}_{2,i}(k-1)+\bm{K}(k)\left[Y_i(k) -\hat{Y}_i(k)\right].
\end{equation*}
Then, subsequent control design can be applicable as well.
\end{remark}\vspace{-2mm}

\subsection{Anti-Synchronization Based Anti-Target Control} \vspace{-2mm}\label{sec: asads}
From the TSE, we can estimate the inter-target distances as $\hat{d}^{12}=\|\bbp_1-\hat{\bbp}_2\|$, $\hat{d}_1^{2}=\|\bp_1-\hat{\bbp}_2\|$ and $\hat{d}_2^{2}=\|\bp_1-\hat{\bbp}_2\|$.
We first define some of the following terms:
\begin{align*}
    &\a, \B \in \R^{+},\ \a \in (-\B^{-1}, 0),\ \bar{\alpha} \triangleq (\alpha-1),\\
    &\D\zeta_i(k) \triangleq (-1)^i(\zeta(k+1)-\a \zeta(k)),i\in\{1,2\}.\ 
    % \D\zeta^{(i)} \triangleq (-1)^i\D\zeta,\ i \in \{1, 2\}.
\end{align*}
% To prevent a protected target from the threats posed by the non-cooperative hostile target,
Hence, the ASATCs can be designed in the following three scenarios.\vspace{-2mm}

\textbf{Case 1:} When $\hat{d}^{12}\in\Omega_1$, set $r(k) \equiv r_1$,\vspace{-2mm}
\begin{align}
\bu_{i}
=2t^{-2}
\left[
    \bar{\alpha}\bg^1\bq_{i}^{1} +\D\zeta_i
\right]
+2t^{-1}(\bbv_1-\b{v}_i),\label{eq12-1}
% \\
% \\
% \bu_{2}
% &=2t^{-2}
% \left[
%     \bar{\alpha}\bg(\hat{d}_1^1,\hat{d}_2^1)\bq_{2}^{1}-\D\zeta^{(i)}
% \right]
% +2t^{-1}(\bbv_1-\b{v}_2),\label{eq12-2}
\end{align}
where $r_1$ is a given constant. The function 
$\bg^j, j\in\{1,2\}$ is defined as $\bg^j\triangleq\bg(\hat{d}_1^j,\hat{d}_2^j)=\frac{1}{\beta}\big(\frac{U(\beta-1)}{\max\{U, (\hat{d}_1^j+\hat{d}_2^j)/2\}}+1\big)$, where $U \in \R$ and $                                   \beta >1$, $\beta$ is a given positive integer. Based on the above, we ensure $\bg^j\in (\beta^{-1},1]$.  The role of each term in \eqref{eq12-1} is further clarified in Remark \ref{remark_controller}. In this case, $\hat{d}_1^1=\|\bq_{1}^{1}\|$ and $\hat{d}_2^1=\|\bq_{2}^{1}\|$. \vspace{-2mm}

\textbf{Case 2:} When $\hat{d}^{12}\in\Omega_2$, set $r(k) \equiv r_1$,\vspace{-2mm}
\begin{align}
\bu_{i}
=&
2t^{-2}
\left[
    \bar{\alpha}\bg^2\hat{\bq}_{i}^2+\D\zeta_i
\right]
+2t^{-1}(\hat{\bbv}_2-\b{v}_i),
\label{eq12-3}
% \\
% \bu_{2}
% =&
% 2t^{-2}
% \left[
%     \bar{\alpha}\bg(\hat{d}_1^2,\hat{d}_2^2)\hat{\bq}_{2}^2-\D\zeta^{(i)}
% \right]
% +2t^{-1}(\hat{\bbv}_2-\b{v}_2),\label{eq12-4}
\end{align}
where $\hat{\bq}_{i}^2, i\in \{1, 2\}$ is the estimated relative position between guardian $i$ and the hostile target, i.e. $\hat{\bq}_{i}^2=\bp_i-\hat{\bbp}_2$.\vspace{-2mm}  

\textbf{Case 3:} When $\hat{d}^{12}\in\Omega_3$, set $r(k) = r_k$,\vspace{-2mm}
\begin{align}
\bu_{i}
=&
2t^{-2}
\left[
    \bar{\alpha}\bg^2\hat{\bq}_{i}^2+\D\zeta_i
\right]
+2t^{-1}(\hat{\bbv}_2-\b{v}_i),
\label{eq12-5}
% \\
% \bu_{2}
% =&
% 2t^{-2}
% \left[
%     \bar{\alpha}\bg(\hat{d}_1^2,\hat{d}_2^2)\hat{\bq}_{2}^2-\D\zeta^{(i)}
% \right]
% +2t^{-1}(\hat{\bbv}_2-\b{v}_2),\label{eq12-6}
\end{align}
where $r_{k+1}=r_{k}-(r_1-r_c)/t_\text{in}$ with the user-defined interception time $ t_\text{in}$. The initial encirclement radius of the guardian when entering the takedown (or capture) zone $\Omega_3$ is defined as $r_0$, and $r_0=r_1$.\vspace{-2mm}

\begin{remark} \label{remark_controller}
The ASATCs in \eqref{eq12-1}-\eqref{eq12-5} can be divided into three parts, e.g., in \eqref{eq12-1}, $\bu_{1}=\bu_{cir}+\bu_{tra}+\bu_{com}$, where the encirclement acceleration is given by $\bu_{cir}=\frac{2\alpha}{t^2}\left[\bg^1\bq_{1}^1+\bzeta(k)\right]$, the tracking acceleration by $\bu_{tra}=\frac{-2}{t^2}\left[\bg^1\bq_{1}^1+\bzeta(k+1)\right]$, and the compensation acceleration by $\bu_{com}=2t^{-1}(\bbv_1-\b{v}_1)$. Based on this design, we can ensure that guardians surround the target on a trajectory of the shape $\bzeta(k)$, as stated in Problem \ref{problem}.
\end{remark}\vspace{-2mm}

\begin{remark} 
To account for the maximum flight speed of a real-world guardian, the function $\bg^j$ is employed to constrain the magnitude of $\bu_i$. The function $\bg^j$, which depends on two distance measurements between the two guardians and the target, is set to ensure that the tracking control does not overwhelm the encirclement control, which may cause the loss of persistent excitation, as will be seen from Lemma 2 later.
Moreover, we must ensure that $\bg^j$ does not approach zero. When $\bg^j\rightarrow0$, $\tilde{\alpha}_j=\bg(\hat{d}_1^j,\hat{d}_2^j)(\alpha-1)+1\rightarrow1, j\in \{1, 2\}$ in the corresponding encirclement error systems \eqref{eq20} and \eqref{eq21}, and the systems will not converge.
\end{remark}\vspace{-2mm}

\subsection{Convergence Analysis}\vspace{-2mm}

The TSE in \eqref{eq9-1} is a Kalman Filter for LTV system with the system matrices $A(k)$ and $B(k)$ are time-invariant, i.e. $A(k) \equiv A$ and $B(k) \equiv B$, while the observation matrix $C(k) = [\bq_{12}^\top\ \bm{0}]$ is time-varying. Before further analysis on stability and converge of the overall system against the objectives in Sec.\ref{sec: problem fomulation}, we shall recall some properties of this class of estimators:\vspace{-2mm}

\begin{definition}
\label{Uniform Observability} \cite{deyst1968conditions} (Uniform observability) \label{Uniform observability)}
\textit{The matrix pair $\{A(k), C(k)\}$, where $A(k), C(k)\in \R^{3\times3}$ are system and output matrices, respectively, is uniformly completely observable if $\exists$ $\hat{a}_{S2}, \check{a}_{S2} \in \R^{+}$, $M \in \Z^+$ and $\Gamma_2\in \R^{+}$ such that for all $k \geq 0$,
\begin{equation}\label{eq: A C PE}
\begin{split}
\hat{a}_{S2}I\leq\sum^{k+M-1}_{m=k}
&(\mathcal{A}^m)^\top C(m)^\top \Gamma_2 ^{-1}C(m)\mathcal{A}^m\leq\check{a}_{S2}I,
\end{split}
\end{equation}
where $\mathcal{A}^m=A^{-(k+M-1-m)}$, and $\mathcal{A}^{k+M-1}=I$.}
\end{definition} \vspace{-2mm}

\begin{lemma}\cite{haring2020stability} (Bounds of the Kalman filter variance) \label{Bounds of the Kalman filter variance}
Given the following assumptions:\vspace{-4mm}
\begin{enumerate}
\item The matrices $A(k)$, $B(k)$, and $C(k)$ satisfy: $\hat{c}_A I\leq A(k)A(k)^\top \leq \check{c}_AI$, $B(k)B(k)^\top \leq \check{c}_B I$ and
$C(k)^\top C(k) \leq \check{c}_C I$ for some $\hat{c}_A,\check{c}_A,\check{c}_B,\check{c}_C \in \R^{+}$ and all $k\geq0$.
\item The noise variance of the process $W_2$ and the measurement noise variance $\Gamma_2$ satisfy: $
\hat{c}_W I \leq W_2\leq \check{c}_W I$,$\hat{c}_\Gamma I\leq\Gamma_2\leq \check{c}_\Gamma I$
for some $\hat{c}_W,\check{c}_W, \hat{c}_\Gamma,\check{c}_\Gamma \in \R^{+}$ and all $k\geq0$.
\item The system is uniformly observable and controllable.
\end{enumerate}\vspace{-4mm}
Then, the Kalman filter variance $\bm{G}(k)$ in \eqref{eq9-4} has upper and lower bounds, i.e. $\exists$ $\hat{\bm{G}}_1, \check{\bm{G}}_1 \in \R^{+}$ such that: $\hat{\bm{G}}_1I\leq\bm{G}^{-1}\leq\check{\bm{G}}_1I$ for all $k\geq0$.
\end{lemma} \vspace{-2mm}

Secondly, to ensure the observability of the hostile target, the following lemmas are provided.\vspace{-2mm}

\begin{lemma}(Persistently excitation)\label{lem: position_Persistently_exciting}
Under ASATCs in \eqref{eq12-1}-\eqref{eq12-5}, the relative position $\bq_{12}(k)$ is PE, i.e. $\exists$ $\hat{a}_{\bq}, \check{a}_{\bq} \in \R^{+} $ and $N \in Z^+$ such that for all $k$: \vspace{-2mm}
\begin{equation}\label{eq15-1}
\begin{split}
\hat{a}_{\bq}I\leq\tilde{S}\triangleq \sum_{m=k}^{k+N-1}\bq_{12}(m) \bq_{12}^\top(m)\leq\check{a}_{\bq}I.
\end{split}
\end{equation}
\vspace{-0.5cm}

\end{lemma}
\textbf{Proof.} Considering the movement of the two guardians in \eqref{eq: dynamic} with the ASATCs in \eqref{eq12-1}-\eqref{eq12-5} and the relative position $\bq_{12}$ in \eqref{eq3}, we can derive the dynamic model of the relative position $\bq_{12}$ as follows:
\begin{equation}\label{eq15}
\bq_{12}(k+1)
=\tilde{\alpha}_j(k)\bq_{12}(k)
+2\left[
\alpha\bzeta(k)-\bzeta(k+1)\right],
\end{equation}
where $\tilde{\alpha}_j(k)=\bg(\hat{d}_j^2,\hat{d}_j^2)(\alpha-1)+1$.\vspace{-2mm}

We first show that $S_1(k) \triangleq \alpha\bzeta(k)-\bzeta(k+1)$ is PE,
which is equivalent to proving that $\hat{a}_{S_1} \leq \tilde{S_1} \leq \check{a}_{S_1}$ with $\tilde{S_1}\triangleq \sum_{m=k}^{k+N-1}\bfx^\top S_1(m)(S_1(m))^\top \bfx$ for some finite $\hat{a}_{S_1},\ \check{a}_{S_1} \in \R^{+}$, $N \in \Z^{+}$ and all unit vectors $\bfx$.
First, observe that for all $k>0$ and any unit vector $\bfx$,\vspace{-2mm}
\begin{equation*}
\begin{split}
\tilde{S_1}
=
&\sum_{m=k}^{k+N-1}\alpha^2\left[\bfx^\top\bzeta(m)\right]^2
+\sum_{m=k}^{k+N-1}\left[\bfx^\top\bzeta(m+1)\right]^2\\
&-\sum_{m=k}^{k+N-1}2\alpha\bfx^\top\bzeta(m)\bfx^\top\bzeta(m+1).
\end{split}
\end{equation*}\vspace{-0.5cm}

Utilizing the Cauchy-Schwartz inequality, we have 
\begin{equation*}
\begin{split}
% -&\sqrt{\sum_{m_1=k}^{k+N-1}
% \left[\bfx^\top\bzeta(m)\right]^2\sum_{m=k}^{k+N-1}\left[\bfx^\top\bzeta(m+1)\right]^2}
% \\
% \leq
&\sum_{m=k}^{k+N-1}\bfx^\top\bzeta(m)\bfx^\top\bzeta(m+1)\\
&\leq\sqrt{\sum_{m_1=k}^{k+N-1}(\bfx^\top\bzeta(m_1))^2\sum_{m=k}^{k+N-1}\left[\bfx^\top\bzeta(m+1)\right]^2}.
\end{split}
\end{equation*}
Then, by using Assumption \ref{Persistently_exciting_shape}, we can quickly obtain $\sum_{m=k}^{k+N-1}\bfx^\top\bzeta(m)\bfx^\top\bzeta(m+1) \leq \check{a}_{\bzeta}$, and therefore:
\begin{equation*}
\begin{split}
&\hat{a}_{S1}\leq\sum_{m=k}^{k+N-1}
\left[\bfx^\top(\alpha\bzeta(m)-\bzeta(m+1))\right]^2\leq\check{a}_{S1},
\end{split}
\end{equation*}
where $\hat{a}_{S1} = \alpha^2\hat{a}_\zeta-2\alpha\check{a}_\zeta+2\hat{a}_\zeta$ and $\check{a}_{S1} = (\alpha\sqrt{\check{a}_\zeta}+\sqrt{\check{a}_\zeta})^2$.\vspace{-2mm}

Considering $\check{a}_\zeta>0$, $\hat{a}_\zeta>0$ and $\alpha<0$, $\hat{a}_{S1} > 0$, we can conclude that $2\alpha\bzeta(k)-2\bzeta(k+1)$ is PE.\vspace{-2mm}

Furthermore, from \eqref{eq15}, we have $\bq_{12}
=\prod_{m_1=0}^{k}\tilde{\alpha}_j(m_1)\\ \bq_{12}(0)+\sum_{m=0}^{k-1}\prod_{m_1=m}^{k-1}\tilde{\alpha}_j(k-1-m_1)2 S_1(m)$. As $\alpha \in (-\beta^{-1}, 0)$, $\bg(\hat{d}_1^j,\hat{d}_2^j) \in (\beta^{-1},1]$ and $\beta >1$, we have $0<|\tilde{\alpha}_j(k)|< \max\{\beta^{-1}, 1-\beta^{-1}\}<1$ for all $k$, and $\bfx^\top\tilde{S}\bfx\geq \sum_{m=k}^{k+N-1}\big( \sum_{m_2=0}^{m-1}\prod_{m_1=m_2}^{m-1}\tilde{\alpha}_j(m-1-m_1)2 \bfx^\top S_1(m_2)\big)^2\geq\sum_{m=k}^{k+N-1}\big(2 \bfx^\top S_1(m)\big)^2$. 
Considering $S_1$ is PE, $\{\bq_{12}\}$ will also become PE.\vspace{-2mm}

\begin{lemma}\label{Uniform Observability_1}
Under ASATCs in \eqref{eq12-1}-\eqref{eq12-5}, the hostile target system \eqref{eq2} with noisy distance measurement is uniformly observable.
\end{lemma}\vspace{-2mm}
\textbf{Proof.}
According to Definition \ref{Uniform Observability}, we need to demonstrate that $S_2(k) \triangleq (\mathcal{A}^{k})^\top C(k)^\top\Gamma_2^{-1}C(k)\mathcal{A}^{k}$ is PE, i.e. $\exists$ $\hat{a}_{S2}, \check{a}_{S2}, \ell
\in \R^{+} $ and $\exists$ $M\geq \ell
N \in \Z^+$ such that
$\hat{a}_{S2} \leq \tilde{S_2} \leq \check{a}_{S2}$ where $\tilde{S_2}\triangleq\sum^{k+M-1}_{m=k}\bfx^\top S_2(m)\bfx$.\vspace{-2mm}

Recalling the expression of $A$ in \eqref{eq: dynamic}, the following can be obtained,\vspace{-2mm}
\begin{equation*}
\begin{split}
\mathcal{A}^{m}=\left[
    \begin{array}{cc}
      I &\hbar_{m}I \\
      \bm{0} & I \\
    \end{array}
      \right],
\end{split}
\begin{split}
C^\top\Gamma_2^{-1}C=\left[
                \begin{array}{cc} \Gamma_2^{-1}\bq_{12}\bq_{12}^\top&\bm{0}\\
                 \bm{0}&\bm{0}\\
                \end{array}
              \right],
\end{split}
\end{equation*}
where $\hbar_{m}=-(k+M-1-m)t$.
Based on Assumption \ref{variance bound}, the output noise variance $\Gamma_2$ has strictly positive lower and upper bounds, i.e. $\hat{c}_\Gamma\leq \Gamma_2\leq \check{c}_\Gamma$, where $\hat{c}_\Gamma=\frac{1}{2}\hat{\sigma}$ and $\check{c}_\Gamma=\frac{1}{2}\check{\sigma}$. Furthermore, from Lemma $\ref{lem: position_Persistently_exciting}$, we have that $\tilde{S}\leq \check{a}I$, which means $\exists \check{a}_{S2}$ such that 
$\tilde{S_2} \leq \check{a}_{S2}$ for $\check{a}_{S2} \in \R^{+}$.\vspace{-2mm}

Then, let us denote all unit vectors in $\R^6$ as $\bfx=[\bfx_1,\bfx_2]^\top$ with $\bfx_1, \bfx_2 \in \R^{3}$ and $|\bfx_1|^2+|\bfx_2|^2=1$, hence:\vspace{-2mm}
\begin{equation*}
\begin{split}
\tilde{S_2}
=&\Gamma_2^{-1}\sum^{k+M-1}_{m=k}((\bfx_1+\hbar_{m}\bfx_2)^\top\bq_{12}(m))^2\\
\geq&\Gamma_2^{-1}\big\{\sum^{\ell
}_{\kappa=1}\sum^{k+\kappa N-1}_{m=k+(\kappa-1) N}((\bfx_1+\hbar_{m}\bfx_2)^\top\bq_{12}(m))^2
\big\},
\end{split}
\end{equation*}

For each $\kappa$, $\hbar_{m}\in\{-(M-\kappa N)t,\ldots,-(M-(\kappa-1)N-1)t\}$. Considering $\tilde{S}\triangleq\sum^{k+\kappa N-1}_{m=k+(\kappa-1) N}\bq_{12}(m)(\bq_{12}(m))^\top$ from Lemma $\ref{lem: position_Persistently_exciting}$, we have: 
\begin{equation*}
\begin{split}
\tilde{S_2}
\geq&\Gamma_2^{-1}\lambda\{\tilde{S}\}\big\{\ell|\bfx_1|^2+c_{M1}t^2|\bfx_2|^2-2c_{M2}t|\bfx_1||\bfx_2|\big\},
\end{split}
\end{equation*}
where $c_{M1}=\sum^{\ell}_{\kappa=1}(M-\kappa N)^2=\ell M^2-\ell(\ell+1)MN+\frac{\ell(\ell+1)(2\ell+1)}{6}N^2>0$ for $M\geq \ell N$ and $c_{M2}=\sum^{\ell}_{\kappa=1}(M-(\kappa-1)N-1)=\ell M-\frac{\ell(\ell-1)}{2}N-\ell$.\vspace{-2mm}

When $|\bfx_1|=0$ or $|\bfx_2|=0$, we can readily conclude that $\hat{a}_{S2} \leq \sum^{k+M-1}_{m=k}\bfx^\top S_2(m)\bfx$ with $\hat{a}_{S2}\in\R^{+}$ based on $\tilde{S}\geq \hat{a}I$ from Lemma $\ref{lem: position_Persistently_exciting}$. Otherwise, when $|\bfx_1|\neq0$ and $|\bfx_2|\neq0$,

we have \vspace{-2mm}
\begin{equation*}
\begin{split}
\tilde{S_2}&\quad \geq \Gamma_2^{-1}\lambda\{\tilde{S}\}\big\{(\sqrt{\ell}|x_1|-\sqrt{c_{M1}}t|x_2|)^2 \dots \\
&\qquad\qquad +2(\sqrt{\ell c_{M1}}-c_{M2})t|\bfx_1||\bfx_2|\big\}\\
&\quad \geq\Gamma_2^{-1}\lambda\{\tilde{S}\}\big\{2(\sqrt{\ell c_{M1}}-c_{M2})t|\bfx_1||\bfx_2|\big\}.
\end{split}
\end{equation*}
There exist $M$, $N$ and sufficiently large $\ell$ such that $\sqrt{\ell c_{M1}}-c_{M2}>0$ and $\tilde{S_2}>0$. For example, if $M=72$, $N=6$ and $\ell=12$, then $\sqrt{\ell c_{M1}}-c_{M2}\approx 11.54$. \vspace{-2mm}

Next, in the same manner as in the proof of [Lemma 2.3 in \cite{hamel2016riccati}], we use the method of contradiction to prove that: $ \exists \hat{a}_{S2}\in \R^{+}$, $\exists M, N, \ell \in \R^{+}$, $\forall \bfx=[\bfx_1,\bfx_2]^\top$ such that $\hat{a}_{S2} \leq \tilde{S_2}$. Denote $\Lambda \triangleq \Gamma_2^{-1}\lambda\{\tilde{S}\}\big\{2(\sqrt{\ell c_{M1}}-c_{M2}) t|\bfx_1||\bfx_2|\big\}$. Here, assume that $ \forall \hat{a}_{S2}\in \R^{+}$,  $ \forall \ell \in \R^{+}$, $\exists \bfx=[\bfx_1,\bfx_2]^\top$ such that $\hat{a}_{S2} > \tilde{S_2}$, which means $\hat{a}_{S2}>\Lambda$. As $\Gamma_2^{-1}$ and $\lambda\{\tilde{S}\}$ are positive and bounded, if $\ell \to \infty$, then $\Lambda \to \infty > \hat{a}_{S2}$, a contradiction. Consequently, we conclude that $ \exists \hat{a}_{S2}\in \R^{+}$, $\exists M, N, \ell \in \R^{+}$, $\forall \bfx$ such that $\hat{a}_{S2} \leq \tilde{S_2}$.\vspace{-2mm}
% where $C_a=((M-N)^2-2(M-1)-1)>0$ and $C_b=(M-N)^2>0$. Based on the properties of the quadratic function, we can derive that $-C_a|x_1|^2+C_b>0$ for $|x_1|\in (0, \sqrt{\frac{C_b}{C_a}})$. Furthermore, since $|\bfx_1|>|\bfx_2|$ and $|\bfx_1|^2+|\bfx_2|^2=1$, it follows that $|\bfx_1|\in (\frac{1}{\sqrt{2}},1)$. Therefore, there can exist $M$ and $N$ such that $\sqrt{\frac{C_b}{C_a}}>1$, which ensures that $-C_a|x_1|^2+C_b>0$. Consequently, $\hat{a}_{S2} \leq \sum^{k+M-1}_{m=k}\bfx^\top S_2(m)\bfx$, where $\hat{a}_{S2}\in \R^{+}$.

% \begin{remark}  
% According to formula \eqref{eq15}, it is evident that the relative position $\bq_{12}$ depends solely on the predefined encirclement vector $\bzeta(k)$, irrespective of which target the guardians will encircle. In other words, whether the guardians can estimate the position of the hostile target is independent of which target the guardians are encircling.
% \end{remark}\vspace{-2mm}

Subsequently, the convergence of the state estimation errors and the AS-based encirclement errors is further analyzed. \vspace{-2mm}

\begin{theorem}
Under TSE in \eqref{eq9-1}, the two guardians can estimate the state of the hostile target, i.e. the state estimation error of the hostile target in the mean square sense converges exponentially to a region bounded by  $\hat{\varepsilon}$.
\end{theorem}\vspace{-2mm}
\textbf{Proof.}
Considering $A$ and $B$ in \eqref{eq: dynamic}, and $C\triangleq\bq_{12}^\top\check{I}_3$ in \eqref{eq8}, we can easily deduce that $\hat{c}_A I\leq AA^\top \leq \check{c}_AI$, $ BB^\top \leq \check{c}_B I$ and $ C^\top C\leq \check{c}_C I$, where $\hat{c}_A= \frac{2+t^2-t\sqrt{4+t^2}}{2}$, $\check{c}_A=\frac{2+t^2+ t\sqrt{4+t^2}}{2}$, $\check{c}_B=1/4t^4+t^2$, and $\check{c}_C=\lambda_{\max}\{\bq_{12}^{T}\bq_{12}\}$ are all positive for $t>0$. \vspace{-2mm}

Furthermore, considering Assumption \ref{variance bound} again, the variance $W_2$ has positive definite lower
and upper bounds, i.e. $\hat{c}_W I \leq W_2\leq \check{c}_W I$ with $\hat{c}_W=\gamma_i \hat{w}_1+(1-\gamma_i)\hat{w}_2$ and $\check{c}_W=\gamma_i \check{w}_1+(1-\gamma_i)\check{w}_2$. Define the controllability matrix as $\mathcal{H}$. Based on the expressions of $A$ and $B$, we have 
\begin{equation*}
\begin{split}
\mathcal{H}=
\left[
                   \begin{array}{cccc}
                     \frac{1}{2}t^2&(\frac{1}{2}+1)t^2&\cdots&(\frac{1}{2}+(M-1))t^2\\
                      t &t&\cdots&t\\
                   \end{array}
                 \right] \otimes I.
\end{split}
\end{equation*}
Since $\mathcal{H}$ has full row rank for some $M$, the hostile target is controllable, i.e.
$\sum^{k+M-1}_{m=k+1}(\mathcal{A}^{m})^{-1}BW_2B^\top(\mathcal{A}^{m})^{-\top}$ has strictly positive definite lower and upper bounds.\vspace{-2mm}

Therefore, based on the above conclusions and considering Lemma \ref{Uniform Observability_1} and Lemma \ref{Bounds of the Kalman filter variance}, we can conclude that $\hat{\bm{G}}_1I\leq\bm{G}^{-1}\leq\check{\bm{G}}_1I$ for all $k\geq0$, which implies that the mean squared estimation error is bounded. Next, we further demonstrate the exponential boundedness of the estimation error.\vspace{-2mm}

Denote the estimation error of the hostile target as $\hat{\be}_{2}\triangleq\bbX_2-\hat{\bbX}_2$. Then, recalling the state model in \eqref{eq2} and the state estimator in \eqref{eq9-1}, the dynamic of $\hat{\be}_{2}$ can be further obtained as
\begin{equation}\label{eq17}
\begin{split}
\hat{\be}_{2}=&\tilde{A}(A\hat{\be}_{2}(k-1)+B\bbu_2(k-1))-KC^\top\bar{\eta},
\end{split}
\end{equation}
where $\tilde{A}=I-K C^\top C$.\vspace{-2mm}

By choosing the Lyapunov function (LF) as $\V_{1}=\hat{\be}_{2}^\top\bm{G}^{-1}\hat{\be}_{2}$, we have
\begin{equation}\label{eq18}
\begin{split}
\hat{\bm{G}}_1\|\hat{\be}_{2}\|^2\leq \V_{1}\leq\check{\bm{G}}_1\|\hat{\be}_{2}\|^2.
\end{split}
\end{equation}\vspace{-10mm}

Furthermore, substituting the formula \eqref{eq17} to $\V_{1}$ and considering $\Expect{\bar{\eta}}=0$ and $\Var\{\bar{\eta}\}=\Gamma_2$, the difference of $\V_{1}$ in the mean square can be obtained as\vspace{-2mm}
\begin{equation*}
\begin{split}
\Expect{\triangle \V_{1}}
=&\Expect{\V_{1}(k+1)-\V_{1}}\\
=&\E\Big((A\hat{\be}_{2}+B\bbu_2)^\top(\tilde{A}(k+1))^\top (\bm{G}(k+1))^{-1} \\
&\times\tilde{A}(k+1)(A\hat{\be}_{2}+B\bbu_2) \Big)\\
&+\Gamma_2C(k+1)(\bm{K}(k+1))^\top(\bm{G}(k+1))^{-1}\\
&\times \bm{K}(k+1)(C(k+1))^\top-\hat{\be}_{2}^\top\bm{G}^{-1}\hat{\be}_{2}.
\end{split}
\end{equation*}\vspace{-8mm}

Based on the definitions of $\bm{G}(k+1)$ and $\bm{G}(k+1|k)$ in Section \ref{sec: tpe}, we have $(\bm{G}(k+1|k))^{-1}=(\bm{G}(k+1))^{-1}\tilde{A}(k+1)$, and therefore, $(\tilde{A}(k+1))^\top (\bm{G}(k+1))^{-1}\tilde{A}(k+1)=(\tilde{A}(k+1))^\top(\bm{G}(k+1|k))^{-1}$ can be obtained. Furthermore, from \eqref{eq9-3}, the formula $(\bm{K}(k+1) (C(k+1))^\top C(k+1))^\top(\bm{G}(k+1|k))^{-1}=(C(k+1))^\top C(k+1)(C(k+1)\\\bm{G}(k+1|k)(C(k+1))^\top+\Gamma_2)^{-1}$ can be derived. Considering $\bm{G}(k+1|k)=A\bm{G}(k+1)A^\top+BW_2B^\top$ and the previously obtained bounds, we can easily deduce that $C(k+1)\bm{G}(k+1|k)(C(k+1))^\top+\Gamma_2\leq\Big( \check{c}_C\big(\check{c}_A\check{\bm{G}}_1+\check{c}_B(\check{c}_W)\big)\hat{c}_\Gamma^{-1}+1\Big)\Gamma_2\triangleq c_G\Gamma_2$, where $c_G>1$. Therefore, based on the definition of $\tilde{A}(k+1)$ in \eqref{eq17}, we have $(\tilde{A}(k+1))^\top (\bm{G}(k+1))^{-1}\tilde{A}(k+1)\leq(\bm{G}(k+1|k))^{-1}-C(k+1))^\top(c_G\Gamma_2)^{-1} C(k+1)\leq\bm{G}(k+1|k))^{-1}$.\vspace{-2mm}

Furthermore, recalling the estimator gain $K$ in \eqref{eq9-3}, we have $\bm{K}(k+1)=\bm{G}(k+1)\Gamma_2^{-1} \leq \frac{1}{\hat{\bm{G}}_1\check{c}_\Gamma} I$. Then, since $(\bm{G}(k+1|k))^{-1}\leq (A\bm{G} A^\top)^{-1}\leq \frac{\check{\bm{G}}_1}{{\hat{c}_A}}I$ and $\Expect{\bbu_2^\top\bbu_2}=\trace\{W_2\}\leq 3\check{c}_W$, we have
\begin{equation}\label{eq18-1}
\begin{split}
&\Expect{\triangle \V_{1}}\\
\leq&-\frac{1}{c_G}\E\Big(\hat{\be}_{2}^\top A^\top (C(k+1))^\top\Gamma_2^{-1} C(k+1) A\hat{\be}_{2}\Big)\\
&+\Expect{(B\bbu_2)^\top(A\bm{G} A^\top)^{-1}(B\bbu_2)}
\\
&+(C(k+1))^\top(\bm{G}(k+1) \Gamma_2^{-1})^\top  C(k+1)\\
\leq&-\frac{1}{c_G}\hat{\be}_{2}^\top A^\top (C(k+1))^\top\Gamma_2^{-1} C(k+1)A\hat{\be}_{2}+\tilde{\varepsilon},
\end{split}
\end{equation}
where $\tilde{\varepsilon}=\frac{3}{\hat{c}_A}\check{\bm{G}}_1\check{c}_B\check{c}_W+\frac{\check{c}_C}{\hat{c}_\Gamma\hat{\bm{G}}_1}$.\vspace{-2mm}

From \eqref{eq17}, we have $\Expect{\hat{\be}_{2}}=\tilde{A}A\hat{\be}_{2}(k-1)$. Furthermore, since $\tilde{A}=\bm{G}(\bm{G}(k|k-1))^{-1}>\bm{0}$ and based on the results of Lemma \ref{lem: position_Persistently_exciting} and Lemma \ref{Uniform Observability_1}, we can draw a conclusion that $\sum^{k+M-1}_{m=k}(\tilde{\mathcal{A}}^{m})^\top(C(m))^\top\Gamma_2^{-1}C(m)\tilde{\mathcal{A}}^{m}\triangleq O$ has strictly positive definite upper and lower bounds, where  $\tilde{\mathcal{A}}^{m}=(\tilde{A}A)^{k+M-1-m}$. Therefore, we can further have \vspace{-2mm}
\begin{equation}\label{eq19}
\begin{split}
&\E\big(\V_{1}(k+M-1)-\V_{1}\big)
\leq-\frac{1}{c_G}\hat{\be}_{2}^\top A^\top OA\hat{\be}_{2}+\tilde{\varepsilon}
\\
&\quad\leq-\frac{\lambda_{\min}\{O\}\hat{c}_A}{c_G\check{\bm{G}}_1}\V_{1}+\tilde{\varepsilon} 
\leq-c_\alpha\V_{1}+\tilde{\varepsilon},
\end{split}
\end{equation}
where $c_\alpha \in(0,1]\cap(0,\frac{\lambda_{\min}\{O\}\hat{c}_A}{c_G\check{\bm{G}}_1}]$ is a constant.\vspace{-2mm}

In view of \eqref{eq18} and \eqref{eq19}, and based on Lemma 2.1 of \cite{reif1999stochastic}, Lemma 1 of \cite{liu2016stochastic} and Appendix A of \cite{nguyen2019persistently}, we can obtain that the mean squared estimation error is exponentially bounded, i.e. $\Expect{\|\hat{\be}_{2}\|^2}\leq \big[(1-c_\alpha)^{\ell}\check{\bm{G}}_1\Expect{\|\hat{\be}_{2}(k-\ell(M-1)\|^2}+\sum_{\imath=0}^{\ell-1}(1-c_\alpha)^{\imath}\tilde{\varepsilon}\big](\hat{\bm{G}}_1)^{-1},\forall k>\ell(M-1),\ell\in\Z^{+}$. Recalling \eqref{eq18-1}, we have $\Expect{\triangle \V_{1}} \leq \tilde{\varepsilon}$, which implies that $\Expect{\|\hat{\be}_{2}(k-\ell(M-1))\|^2}\leq \check{\bm{G}}_1(\hat{\bm{G}}_1)^{-1}\Expect{\|\hat{\be}_{2}(0)\|^2}+(M-2)(\hat{\bm{G}}_1)^{-1}\tilde{\varepsilon}$ for $k-\ell(M-1)\in\{0,\ldots,M-2\}$. In conclusion, $\Expect{\|\hat{\be}_{2}\|^2}\leq \hat{\varepsilon},\forall k>\ell(M-1),\ell\in\Z^{+}$ with $\hat{\varepsilon}=(1-c_\alpha)^{\ell}\check{\bm{G}}_1(\hat{\bm{G}}_1)^{-1}\Expect{\|\hat{\be}_{2}(0)\|^2}+(\hat{\bm{G}}_1)^{-1}\big(\sum_{\imath=0}^{\ell-1}(1-c_\alpha)^{\imath}+(M-2)\big)\tilde{\varepsilon}$.\vspace{-2mm}

\begin{theorem} \label{encirclement error  convergence}
Under the ASATCs in \eqref{eq12-1}, the two guardians can achieve the encirclement of the protected target, i.e. the AS-based encirclement error of the protected target in the mean square sense converges exponentially to a region bounded by $\varepsilon_1$ for the controller gain $\alpha \in (-\beta^{-1}, 0)$.
\end{theorem}\vspace{-2mm}
\textbf{Proof.}
Considering the formulas \eqref{eq5-2} and \eqref{eq5-3} in Problem \ref{problem}, we can define the AS-based encirclement error of the target $j$ for each guardian $i$ as $\bar{\be}_{j}\triangleq\bq_{1}^{j} + \bq_{2}^{j}$.\vspace{-2mm}

For all $k\in\{k|\hat{d}^{12} \in \Omega_1\}$, based on the guardians and protected target models in \eqref{eq: dynamic} and \eqref{eq2}, the following dynamics of $\bar{\be}_{1}$ can be further derived.
\begin{equation}\label{eq20}
\begin{split}
\bar{\be}_{1}(k+1)=&\tilde{\alpha}_1\bar{\be}_{1}+\frac{1}{2}t^2\bbu_1,
\end{split}
\end{equation}
where $\tilde{\alpha}_1=\bg(\hat{d}_1^1,\hat{d}_2^1)(\alpha-1)+1$.\vspace{-2mm}

By choosing the Lyapunov function candidate $\V_{1}=\|\bar{\be}_{1}\|^2$ , the differences of $\V_{1}$ in the mean square can be obtained as follows: $\Expect{\triangle \V_{1}} 
=(\tilde{\alpha}_1^2-1)\V_{1}+t^4\Expect{\bbu_1^T\bbu_1}$. Noting that there exists a positive constant $\tilde{\beta}$ such that $\tilde{\beta}^{-1}\triangleq\max\{\beta^{-1}, 1-\beta^{-1}\}<1$, we have $|\tilde{\alpha}_1|<\tilde{\beta}^{-1}$. Then, considering $\Expect{\bbu_1^T\bbu_1}=\trace\{W_2\}$ with $\hat{c}_W I \leq W_2\leq \check{c}_W I$, $\Expect{\triangle \V_{1}}$ can be re-obtianed as \vspace{-2mm}
\begin{equation*}
\begin{split}
\Expect{\triangle \V_{1}} 
\leq& (\tilde{\beta}^{-2}-1)
\V_{1}+3t^4\check{c}_W.
\end{split}
\end{equation*}
\vspace{-0.75cm}

Then, based on Lemma 2.1 of \cite{reif1999stochastic}, we have that the AS-based encirclement error in the mean square sense converges exponentially to a region bounded by $\varepsilon_1$, i.e. $\Expect{\|\bq_{1}^{1}+\bq_{2}^{1}\|^2}\leq \varepsilon_{1}$, $\forall \ k\in\{k|\hat{d}^{12} \in \Omega_1\}$, where the error bound $\varepsilon_{1}=\tilde{\beta}^{-2k}\|\bq_{1}^{1}(0)+\bq_{2}^{1}(0)\|^2+\sum_{\imath=0}^{k-1}\tilde{\beta}^{-2\imath}(3t^4\check{c}_W)$.\vspace{-2mm}

If there is no constraint on the control input, that is, if $\bg(\hat{d}_1^2,\hat{d}_2^2)=1$, then we have $\tilde{\alpha}=\alpha\in (-\beta^{-1}, 0)$ with constant $\beta >1$ in \eqref{eq20}, and $\Expect{\triangle \V_{1}}$ can be rewritten as
\begin{equation*}
\begin{split}
\Expect{\triangle \V_{1}} 
=&
(\alpha^2-1)\V_{1}+t^4\Expect{\bbu_1^T\bbu_1}\\
\leq&
(1/\beta^2-1)\V_{1}+3t^4\check{c}_W.
\end{split}
\end{equation*}
\vspace{-0.5cm}

In this case, as $k$ approaches infinity, the upper bounds of the encirclement errors are only related to the target acceleration $\bbu_1$ and the sampling time $t$.\vspace{-2mm}

\begin{theorem}
Under the ASATCs in \eqref{eq12-3} and \eqref{eq12-5}, the two guardians can achieve the encirclement of the hostile target, i.e. the AS-based encirclement error of the hostile target in the mean square sense converges exponentially to a region bounded by $\varepsilon_2$ for the controller gain $\alpha \in (-\beta^{-1}, 0)$.
\end{theorem}\vspace{-2mm}
\textbf{Proof.}
When $\hat{d}^{12} \in (\Omega_2 \cup \Omega_3)$, the AS-based encirclement error of the hostile target is defined as $\bar{\be}_{2}\triangleq\bq_{1}^{2}+\bq_{2}^{2}$.
Then, the dynamics of $\bar{\be}_{2}$ can be obtained as
\begin{equation}\label{eq21}
\begin{split}
\bar{\be}_{2}(k+1)=&\tilde{\alpha}_2\bar{\be}_{2}-2(\tilde{\alpha}_2-1)\hat{\be}_2+t^2\bbu_2,
\end{split}
\end{equation}
where $\tilde{\alpha}_2=\bg(\hat{d}_1^2,\hat{d}_2^2)(\alpha-1)+1$.\vspace{-2mm}

Similarly, by choosing the Lyapunov function candidate $\V_{2}=\|\bar{\be}_{2}\|^2$, the difference of $\V_{2}$ in the mean square can be obtained as follows,\vspace{-2mm}
\begin{equation*}
\begin{split}
\Expect{\triangle \V_{2}} 
=& (\tilde{\alpha}_2^2-1)\V_{2}-4\tilde{\alpha}_2(\tilde{\alpha}_2-1)\Expect{\bar{\be}_{2}^\top\hat{\be}_2}\\
&+4(\tilde{\alpha}_2-1)^2\Expect{\|\hat{\be}_2\|^2}+t^4\Expect{\bbu_2^T\bbu_2}.
\end{split}
\end{equation*}\vspace{-2mm}

Assume that there exists $\varrho \in (0,1)$ such that $\frac{\sqrt{1+\varrho}}{\sqrt{1+\varrho}-1}>\beta>\sqrt{1+\varrho}$. Using Young’s inequality $2\|a\|\|b\|\leq \varrho \|a\|^2+\varrho^{-1}\|b\|^2$ yields $-4\tilde{\alpha}_2(\tilde{\alpha}_2-1)\Expect{\bar{\be}_{2}^\top\hat{\be}_2}\leq \varrho\tilde{\alpha}_2^2\Expect{\|\bar{\be}_2\|^2}+4\varrho^{-1}(\tilde{\alpha}_2-1)^2\Expect{\|\hat{\be}_2\|^2}$. Based on $\Expect{\|\hat{\be}_{2}\|^2}\leq \hat{\varepsilon}$ proved in Theorem \ref{encirclement error  convergence} and $|\tilde{\alpha}_2|<\tilde{\beta}^{-1}$ with constant $\tilde{\beta}^{-1}\triangleq\max\{\beta^{-1}, 1-\beta^{-1}\}<1$, we have
\begin{equation*}
\begin{split}
\Expect{\triangle \V_{2}} 
\leq&
((1+\varrho)\tilde{\beta}^{-2}-1)
\V_{2}+\tilde{e}_\varepsilon+3t^4\check{c}_W,
\end{split}
\end{equation*}
where $\tilde{e}_\varepsilon=4(1+\varrho^{-1})(\tilde{\alpha}_2-1)^2 \hat{\varepsilon}$.

Since $0<(\sqrt{1+\varrho}-1)^2<(1+\varrho)\tilde{\beta}^{-2}<(1+\varrho)^{-1}<1$, the AS-based encirclement error $\bar{\be}_{2}$ in the mean square sense converges exponentially to a region bounded by $\varepsilon_2$, i.e. $\Expect{\|\bq_{1}^{2}+\bq_{2}^{2}\|^2}\leq\varepsilon_{2}, \forall k \in \{k|\hat{d}^{12} \in (\Omega_2 \cup \Omega_3)\}$ with $\varepsilon_{2}=((1+\varrho)\tilde{\beta}^{-2})^{k}\|(\bq_{1}^{2})(0)+(\bq_{2}^{2})(0)\|^2+\sum_{\imath=0}^{k-1}((1+\varrho)\tilde{\beta}^{-2})^{\imath}(\tilde{e}_\varepsilon+3t^4\check{c}_W)$.\vspace{-2mm}

\section{Simulations and Experiments}\vspace{-2mm}
This section presents the main simulation and real world experiment results. Full results and visualizations are available at \url{https://youtu.be/CVeIw8pKbh4}. \vspace{-2mm}

\textbf{Numerical simulation:}
A simulation example validates that the designed TSE and ASATC can both perform well. Here, we consider two guardians, one ground cooperative car (protected target) and one hostile target. The acceleration motions of targets are given as $\P\{\pi_1\}=1^{\pi_1}$, $\P\{\pi_2\}=0.95^{\pi_2}0.05^{(1-\pi_2)}$, $\bm{W}_{1,1}=\textbf{diag}([10^{-3}, 10^{-3}, 0])$, $\bm{W}_{1,2}=\textbf{diag}([0.0008, 0.002, 0])$, and $\bm{W}_{2,2}=\textbf{diag}([0.004, 0.01, 10^{-4}])$. \vspace{-2mm}

For distance measurement noise, we assume that $\sigma_{1}=\sigma_{2}=0.1$.
The sampling period $t$ is given as $0.5 s$, the initial positions of all guardians and targets are as follows: \vspace{-2mm}
\begin{equation*}
\begin{split}
&\bp_{1}(0)=[2,\ 2,\ 1]^\top,~\bp_{2}(0)=[0,\ 1.5,\ 0.5]^\top,\\
&\bbp_{1}(0)=[0,\ 0,\ 0]^\top,~\bbp_{2}(0)=[2,\ 12,\ 2]^\top,\\
&\bbv_{1}(0)=[0,\ 0,\ 0]^\top,~\bbv_{2}(0)=[-0.02,\ -0.1,\ 0]^\top,
\end{split}
\end{equation*} \vspace{-2mm}
and the initial velocities of all guardians are zeros.\vspace{-2mm}

\begin{figure}
\centering
  \includegraphics[width=7.5cm]{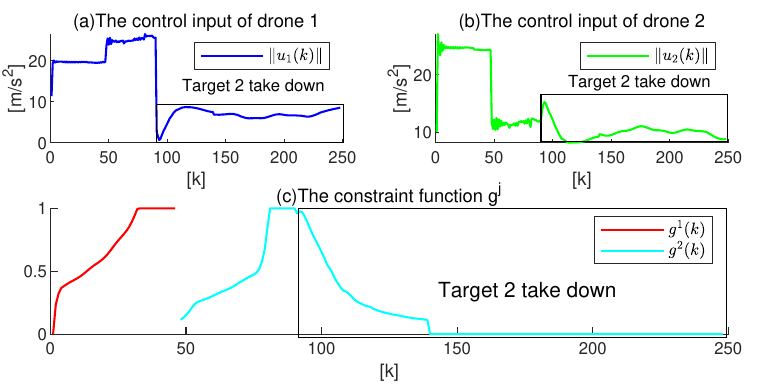}
 \caption{The trajectories of the control acceleration and the constraint function.}
  \label{control_input}
\end{figure} 

\begin{figure}
\centering
  \includegraphics[width=7.5cm]{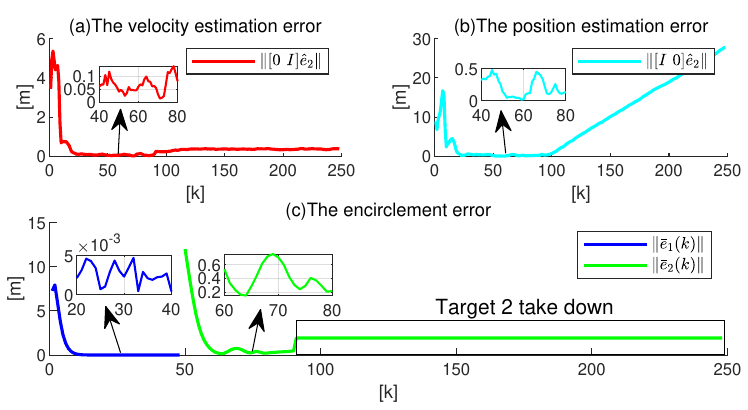}
    \vspace{-10pt}
 \caption{The trajectories of the velocity and position estimation errors of the hostile target, and all AS-based encirclement errors for protected target and hostile target.}
  \label{all_error}
\end{figure}

\begin{figure}
\centering
  \includegraphics[width=8.5cm]{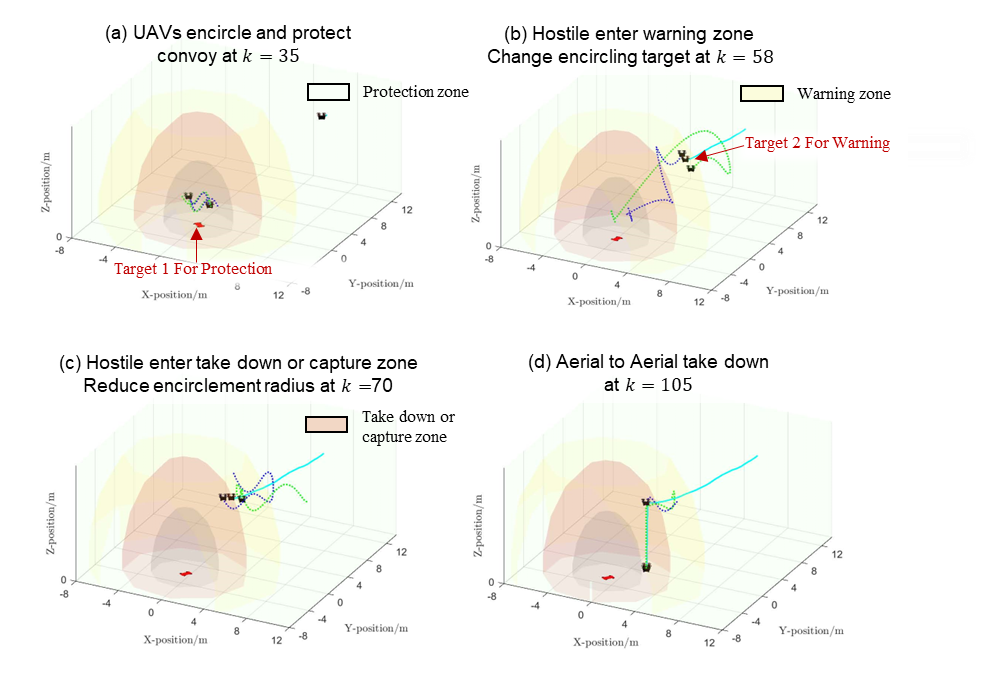}
  \vspace{-25pt}
 \caption{The real-time position points of all targets and guardians.}
  \label{whole_trajectory}
\end{figure}

The encirclement frequency of $\bzeta(k)$ is $\rho=\frac{1}{24}$, and the vertical motion function is designed as $\h(k)=\frac{1}{5}\cos(\frac{1}{8}\pi k)$.  The initial estimation error variance is $\bm{G}(0)=I$. Additionally, we have $r_1=0.9$, $r_c=0.1$ and $t_\text{in}=30$ in the ASATCs. For three
decision zones, we have $l_1=8.5$, $l_2=5.5$ and $l_3=3$. Moreover, guardian 1 and guardian 2 will project the protected target onto the excepted height planes, i.e. $h_1=0.7$.\vspace{-2mm}

Based on Theorem 2 and Theorem 3, $\alpha=-0.1$ is set accordingly. Moreover, $U=1.5$ and $\beta=10$ are set in the acceleration controller of the guardians.\vspace{-2mm}

The simulation results are shown in Fig. \ref{control_input} to Fig. \ref{whole_trajectory}. In Fig. \ref{control_input}, we present the trajectories of the absolute values of the control inputs $\|\bu_1\|$ and $\|\bu_2\|$ for the two guardians, and the trajectories of the constraint function $g^j, j\in\{1,2\}$. From Fig. \ref{control_input}, it can be observed that during the initial stage $(k=1:30)$ and the adaptive transition of the guardians from encircling the protected target 1 to encircling the hostile target 2 during $(k=50:80)$, the constraint function $g^j$ plays a significant role in limiting the magnitude of the controller inputs, resulting in smoother control inputs. The error trajectories in Fig. \ref{all_error} (a) and (b) show that the velocity and position estimation errors for the hostile target decrease with bounds of 0.1 meters and 0.5 meters, respectively, which indicates that AS-based encirclement control contributes to improving the state estimation accuracy for the target. Additionally, from the encirclement error trajectory in Fig. \ref{all_error} (c) and the real-time position points in Fig. \ref{whole_trajectory} (b) and (c), it can be seen that the two guardians can quickly switch from encircling a protected target to encircling a hostile target. For the cooperatively protected target, the encirclement error is close to 0, while for the non-cooperative hostile target, the encirclement error is within 0.6 meters.
\vspace{-2pt}

\begin{figure}
\centering
  \includegraphics[width=8cm]{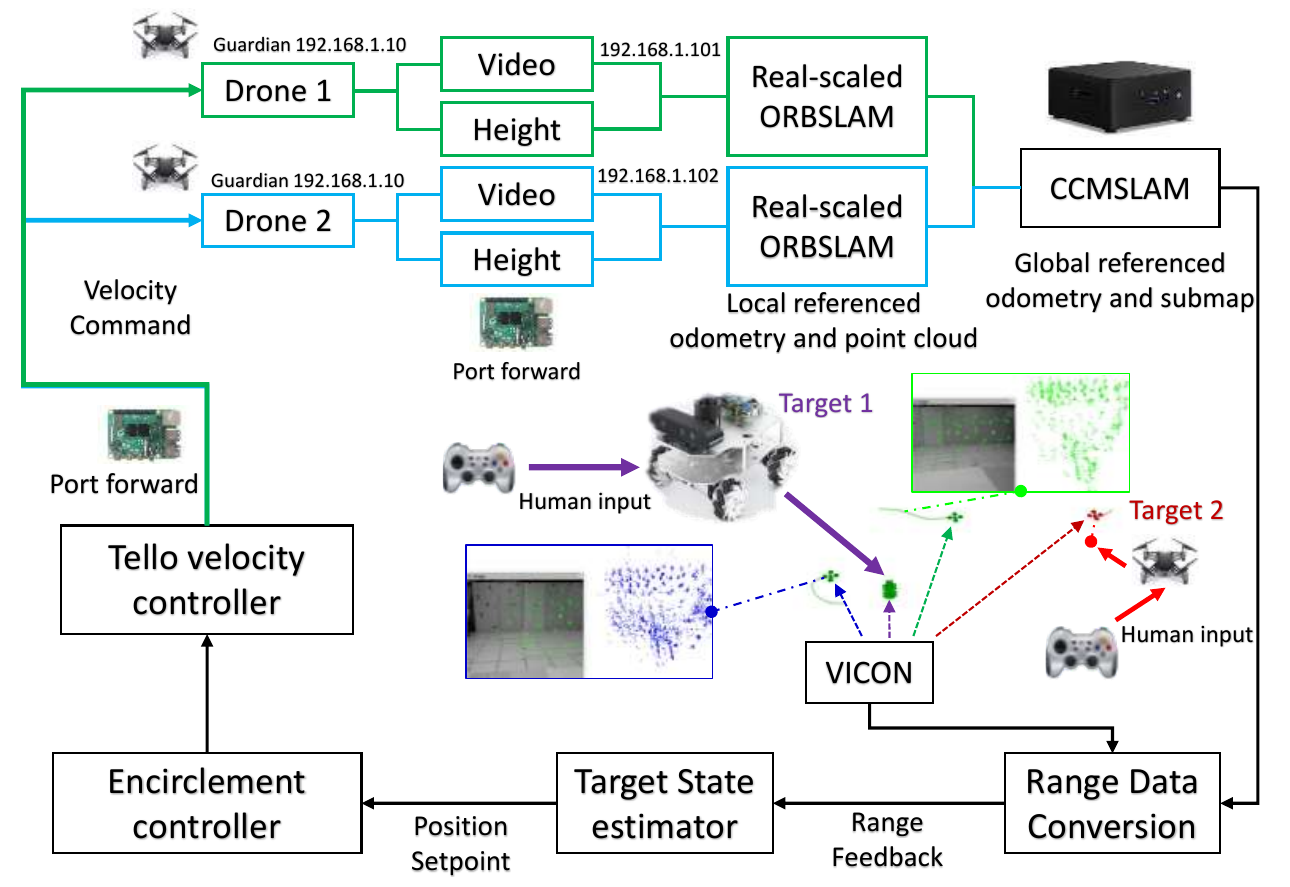}
      \vspace{-10pt}
 \caption{Experiment setup.}
  \label{experiment_setup}
\end{figure}

\textbf{Real-world UAV-based experiment:} To verify the proposed solution, a real-world demonstration is conducted as shown in Fig. \ref{experiment_setup}. The experiment involves one ground patrolling vehicle, two guardian, and one hostile target with the intention of destroying the patrol. \vspace{-2mm}

\begin{figure}
\centering
  \includegraphics[width=7.5cm]{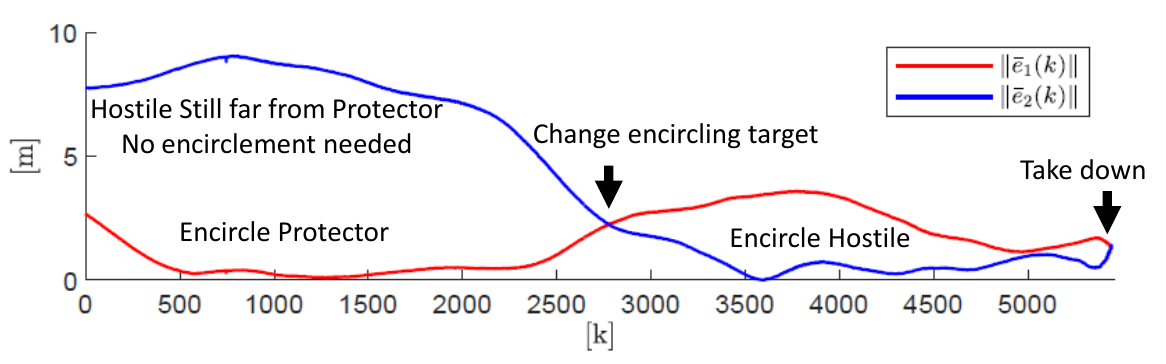}
      \vspace{-10pt}
 \caption{The trajectories of the AS-based encirclement tracking errors $\be_{1}$ and $\be_{2}$ in experiment.}
  \label{Experiment_error}
\end{figure} 

During the experiment of multi-drone flight, the background logging system captured 6,000 state feedback samples from each guardian in real time, achieving an observation rate of 20-25 Hz. Despite bandwidth limitations, the guardians effectively demonstrated their ability to adapt and respond to threats within the short flight duration of less than one minute. Initially, the guardians maintained orbital positioning around the protected target, ensuring its safety. By the 2,500th measurement, they detected the hostile target and repositioned themselves to form a defensive barrier between the threat and the patrol. Between the 3,000th and 5,000th measurements, the guardians shifted their focus to orbiting around the hostile target, carefully monitoring its movements. As the hostile entity advanced, the guardians executed their final maneuver, closing in and neutralizing the threat through direct impact.
%In a 50-second real-world demo, a buffer captured each guardian's state for visualization, marking uninitialized states as unavailable. Out of a theoretical 8500 state feedback instances, bandwidth limits reduced the actual sample count to 6000, yielding an observation rate of 20-25 Hz per guardian. Initially, guardians focused on orbital positioning around the protected target. By the 2500th measurement, they detected a hostile threat and repositioned between the targets. From the 3000th to 5000th measurement, they orbited around the hostile target, eventually closing in and crashing into it as it approached.
\vspace{-2mm}

\section{Conclusions}\vspace{-2mm}
In this research, we have introduced an innovative aerial target encirclement and interception solution that provides robust target protection. This breakthrough is realized by utilizing AS-based control strategies and noisy range measurements. Our findings demonstrate the remarkable efficacy of a specially designed perception-aware controller, allowing the guardian to autonomously neutralize hostile target without necessitating external ground guidance. This autonomous capability holds immense promise, particularly in active war zones, where it has the potential to play a pivotal role in providing surveillance to save lives and enhancing guardian countermeasures.\vspace{-2mm}

\bibliographystyle{unsrt}
\bibliography{autosam}          

\end{document}